%% file: main.tex
\pdfoutput=1
\documentclass[11pt]{article}

\usepackage[preprint]{acl}

\usepackage{times}
\usepackage{latexsym}

\usepackage[T1]{fontenc}
\usepackage[utf8]{inputenc}

\usepackage{microtype}

\usepackage{inconsolata}

\usepackage{graphicx}
\usepackage{natbib}
\usepackage{doi}
\usepackage{tabularx}
\usepackage{booktabs}
\usepackage{amsfonts}
\usepackage{multirow}

\usepackage{color}

\usepackage[toc,page]{appendix}

\title{Language Model Can Do Knowledge Tracing: Simple but Effective Method to Integrate Language Model and Knowledge Tracing Task}
\author{Unggi Lee${^1}$${^,}$${^3}$, Jiyeong Bae${^1}$, Dohee Kim${^1}$\footnotemark[1], Sookbun Lee${^1}$\footnotemark[1], Jaekwon Park${^1}$\footnotemark[1]\\ 
{\bf Taekyung Ahn${^1}$\footnotemark[1], Gunho Lee${^1}$\footnotemark[1], Damji Stratton${^2}$, Hyeoncheol Kim${^3}$\footnotemark[2]} \\
Enuma, Inc.${^1}$, The University of Missouri System${^2}$, Korea University${^3}$ \\
\texttt{\{unggi, jiyoung, dohee, blackdew, jaekwon, taekyung, gunho\}@enuma.com}, \\
\texttt{dhsdfn@umsystem.edu, harrykim@korea.ac.kr}
}

\begin{document}
\maketitle

\footnotetext[1]{Equal contribution}

\input{0_abstract}
\input{1_introduction}

\input{2_related_work}
\input{3_method}

\input{4_result}

\input{5_conclusion}

% Bibliography entries for the entire Anthology, followed by custom entries
%\bibliography{anthology,custom}
% Custom bibliography entries only
% \bibliography{latex/main}
\bibliography{main.bbl}

\end{document}

%% file: 0_abstract.tex
\begin{abstract}
Knowledge Tracing (KT) is a critical task in online learning for modeling student knowledge over time. Despite the success of deep learning-based KT models, which rely on sequences of numbers as data, most existing approaches fail to leverage the rich semantic information in the text of questions and concepts. This paper proposes Language model-based Knowledge Tracing (LKT), a novel framework that integrates pre-trained language models (PLMs) with KT methods. By leveraging the power of language models to capture semantic representations, LKT effectively incorporates textual information and significantly outperforms previous KT models on large benchmark datasets. Moreover, we demonstrate that LKT can effectively address the cold-start problem in KT by leveraging the semantic knowledge captured by PLMs. Interpretability of LKT is enhanced compared to traditional KT models due to its use of text-rich data. We conducted the local interpretable model-agnostic explanation technique and analysis of attention scores to interpret the model performance further. Our work highlights the potential of integrating PLMs with KT and paves the way for future research in KT domain.
\end{abstract}

%% file: 1_introduction.tex
\section{Introduction}

The COVID-19 pandemic has accelerated the adoption of online learning, leading to a significant increase in the number of students participating in digital education platforms \cite{dhawan2020online, leo2021offline}. As online learning continues to expand, the importance of practical tools for assessing and supporting student learning has become increasingly evident \cite{gikandi2011online, ayu2020online}.

\begin{figure}[hbt!]
    \centering
    \includegraphics[width=1\linewidth]{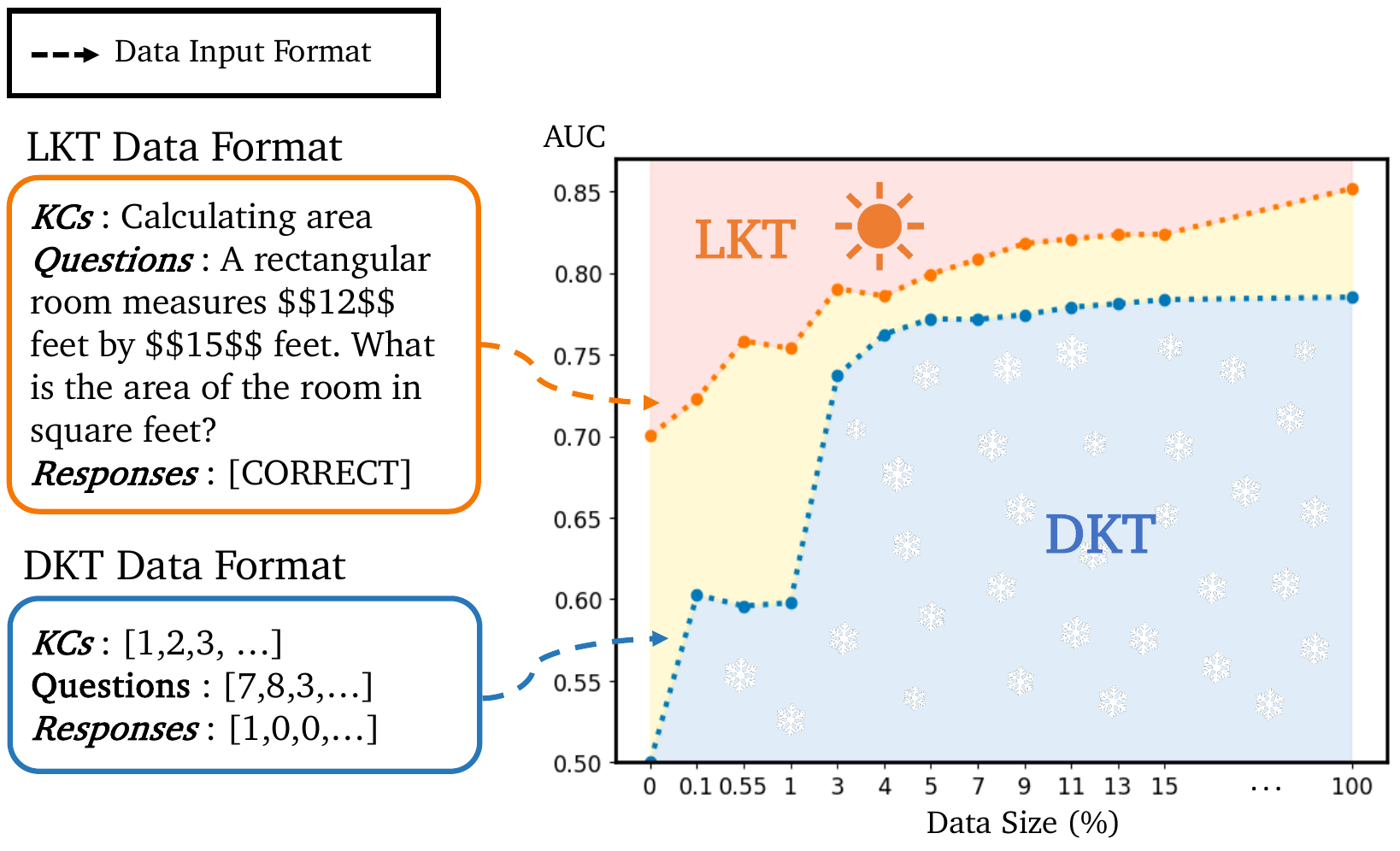}
    \caption{
    Comparison of LKT and DKT on XES3G5M-T dataset. LKT, using RoBERTa with text data, outperforms DKT in both cold start and final AUC performance by leveraging rich text-based semantic information, unlike DKT's numerical sequences. The x-axis shows the proportion of the dataset used for cold start, and the y-axis represents AUC performance.
    }
    \label{fig:init_cold}
\end{figure}

One of the critical components of many online learning platforms is using questions for the formative assessment of student's knowledge and to enhance learning outcomes \cite{ogange2018student}. By utilizing data from these assessments, Knowledge Tracing (KT) models can predict students' knowledge states regarding specific Knowledge Concepts (KCs) and individual question items \cite{corbett1994knowledge}. With many students engaging in online learning, improving the performance of KT models can benefit millions of learners worldwide \cite{shen2024survey, abdelrahman2023knowledge, song2022survey}.

Starting with Bayesian Knowledge Tracing (BKT) \cite{corbett1994knowledge}, the KT domain has evolved to incorporate deep learning techniques, such as Deep Knowledge Tracing (DKT) \cite{piech2015deep}. However, the development of deep learning-based KT models has not kept pace with the rapid advancements in other domains, such as Natural Language Processing (NLP) and Computer Vision (CV).

% Main Image of this paper
\begin{figure*}[hbt!]
    \centering
    \includegraphics[width=0.9\linewidth]{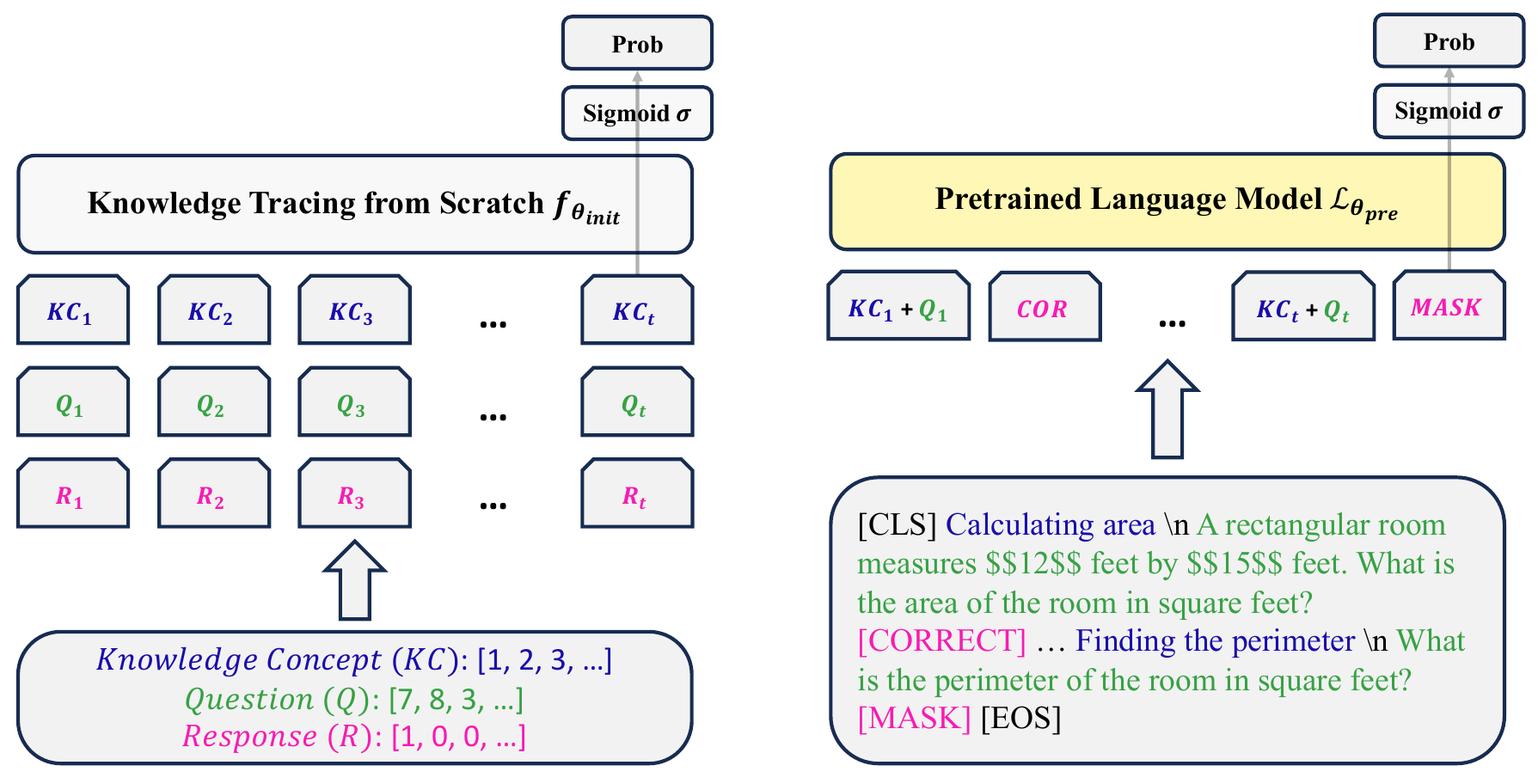}
    \caption{
    The comparison between DKT (\textit{Left}) and LKT (\textit{Right}). LKT uses encoder-based pre-trained LMs ($\mathcal{L}_{\theta_{pre}}$), while DKT models are trained from scratch ($f_{\theta_{init}}$). Data formats differ: DKT uses sequences of numbers (KCs, questions, responses), whereas LKT uses text. The \textit{Bottom} shows interaction data from one student. In LKT, interactions are enclosed by $[CLS]$ and $[EOS]$ tokens, separating KCs and questions. Correctness is indicated by $[CORRECT]$, $[INCORRECT]$, and $[MASK]$ tokens. LKT models predict correctness at the $[MASK]$ position, with 15\% of $[CORRECT]$ or $[INCORRECT]$ replaced by $[MASK]$, inspired by BERT \cite{devlin2018bert}.
    }
    \label{fig:whole_arch}
\end{figure*}

Several limitations hinder the progress of current KT models. First, most KT models rely on sequences of numerical representations for KCs and questions, failing to utilize the rich semantic information contained in the text \cite{liu2019ekt, su2018exercise}. Although a few KT models \cite{su2018exercise, liu2019ekt, jung2023language} have attempted to incorporate text, they primarily use it as an auxiliary means to enhance the model while still relying on numerical sequences as the primary source of model training. This approach is not instinctive, as natural language primarily conveys knowledge \cite{khurana2023natural, lee2022contrastive, liu2019ekt}.

Second, many KT models need help leveraging pre-trained models because they are often tailored to online learning platforms, making it difficult to adapt quickly to other domains, leading to the cold start problem \cite{zhao2020cold}. In contrast, Pre-trained Language Models (PLMs) are more versatile and can be applied to various text-based tasks across different domains, as the text itself serves as a medium for transferring knowledge \cite{devlin2018bert}.

Third, the interpretability of KT models is limited by their reliance on numerical sequences, which lack semantic meaning, unlike NLP models that utilize human-readable text and apply Explainable AI (XAI) techniques \cite{ribeiro2016should, lundberg2017unified}.

Finally, the current state of KT research is far from the mainstream deep learning community, leading to a lack of interest from researchers working on state-of-the-art deep learning techniques \cite{shen2024survey, abdelrahman2023knowledge}. This is evident when comparing the citation scores of DKT \cite{piech2015deep} and other research areas. This disconnection can slow down progress in the KT research area, potentially impacting human learning and future generations significantly.

To address these limitations, we propose a novel Language model-based Knowledge Tracing (LKT) framework, which integrates encoder-based PLMs with KT methods. By leveraging the power of PLMs, such as BERT \cite{devlin2018bert} and RoBERTa \cite{liu2019roberta}, LKT effectively incorporates semantic information from the text of questions and concepts, resulting in significant improvements in KT performance. Our work aims to bridge the gap between KT research and state-of-the-art deep learning techniques, creating new possibilities for advancing the field of KT and ultimately benefiting learners worldwide. The main contributions of our work are as follows:
\begin{itemize}
\item We propose a novel LKT framework that integrates PLMs with KT tasks, enabling accurate predictions of student performance on new questions and concepts even with limited data by leveraging rich semantic information.
\item The LKT framework provides insights into which parts of knowledge concepts and questions affect student performance for learning scientists and educational researchers.
\end{itemize}

%% file: 2_related_work.tex
\section{Literature Review}

\subsection{Knowledge Tracing}

KT is a critical component of Intelligent Tutoring Systems (ITS) that monitors and predicts the development of students' knowledge over time by examining their interactions with instructional content, particularly their responses to questions \cite{piech2015deep, abdelrahman2023knowledge}. From a machine learning perspective, KT is often viewed as a sequence prediction task aimed at estimating the probability that a student correctly answers an upcoming question based on their previous interactions \cite{abdelrahman2023knowledge, lee2022monacobert}.

Deep learning has significantly advanced KT research, with various approaches categorized into six key domains: sequence modeling, memory-augmented models, attentive models, graph-based models, text-aware models, and Forgetting \cite{abdelrahman2023knowledge, piech2015deep, zhang2017dynamic, pandey2019self, ghosh2020context, nakagawa2019graph, su2018exercise, liu2019ekt, liu2021improving}.

These advancements in KT are essential for tailoring instructional materials to meet individual students' needs and implementing targeted strategies to enhance learning outcomes \cite{zhang2017dynamic, abdelrahman2023knowledge, ghosh2020context, lee2022contrastive}.

\subsection{Limitations of Previous Knowledge Tracing Research}

Despite their limitations, KT models have been updated and are showing promising performance \cite{shen2024survey}. Most KT models represent students' interactions with a sequence of numbers, including KC IDs, question IDs. and student responses. However, they often neglect other valuable features like textual content, images, and activity logs within the interaction data. \cite{su2018exercise, liu2019ekt}. This suggests that with further development, KT models could overcome these limitations and offer even more robust performance.

First, KT models struggle to understand the semantic meaning of KCs and questions. Most KT models are mainly trained in KC IDs and question IDs, which consist of number of sequence, they only learn the patterns of students' interaction sequences without understanding what these KCs and questions mean \cite{su2018exercise, liu2019ekt}. Meanwhile, human teachers can understand the semantic meaning of KCs and questions by reading the text \cite{abdelrahman2023knowledge}. Therefore, KT models are trained using an unnatural approach to solve the task.

Second, the cold start problem is a latent issue in KT models \cite{zhao2020cold, das2021new}. KT models use KC and question IDs to train for the target ITS or online learning platform in its present state. If the target ITS or platform adds a new KC or question, the KT models must be retrained from scratch because there are no links between old and new IDs. Moreover, if the target ITS or platform is new, KT models must also be retrained from scratch. A few research studies have tried to explore the self-supervised learning method using pre-training and fine-tuning \cite{liu2021improving, su2018exercise, liu2019ekt, wang2024pre}, but these approaches use only pre-train embeddings and those features are supportive. Therefore, it is necessary to employ models that are pre-trained on main data in an unsupervised manner, similar to the approaches used in the NLP and CV domains.

Third, the interpretability of KT models is limited due to their reliance on numerical sequences that need more semantic meaning. Unlike NLP models that utilize human-readable text and apply Explainable AI (XAI) techniques \cite{ribeiro2016should, lundberg2017unified}, KT models struggle to provide clear explanations for their predictions \cite{li2024explainable}. This lack of interpretability hinders the adoption of KT models in real-world educational settings, as educators and stakeholders require a deep understanding of the factors influencing student performance to make informed decisions and interventions.

%% file: 3_method.tex
\section{Method}

\subsection{Problem Definition}

Each student's learning progress is documented in the DKT setting through a sequence of question-response pairs over time. For student $i$ at time step $t$, the record includes the question they answered, their topic, and whether their answer was correct or incorrect. This is denoted as a tuple $(q_{i}^{t}, c_{i}^{t}, r_{i}^{t})$, where $q_{i}^{t} \in \mathbb{N}_{+}$ is the question index, $c_{i}^{t} \in \mathbb{N}_{+}$ is the topic index, and $r_{i}^{t} \in {0, 1}$ represents the response, with 1 indicating a correct answer. Hence, a record like $(q_{i}^{t}, c_{i}^{t}, 1)$ indicates that the student $i$ correctly answered the question $q_{i}^{t}$ on the topic $c_{i}^{t}$ at time $t$.

Meanwhile, the LKT setting is different. To transform the numerical IDs $q_{i}^{t}$ and $c_{i}^{t}$ into textual features, one can employ techniques like embedding or lookup tables where each ID is mapped to a specific textual description or feature vector. For example, $q_{i}^{t}$ could correspond to the actual text of the question, and $c_{i}^{t}$ could correspond to a description of the topic.

\subsection{Language Model based-Knowledge Tracing}

Masked Language Models (MLMs), such as BERT, employ a special token $[MASK]$, which the model must predict during the pre-training stage to learn the meaning of the language surrounding the $[MASK]$ token. In sequence classification tasks, MLMs typically use the special token $[CLS]$ to capture the meaning of the entire sequence, which starts with $[CLS]$ and ends with $[EOS]$. The MLM is only required to predict the label using the representation of the $[CLS]$ token. MLMs are also used in token classification tasks, where the model predicts the label for each text token. This approach is commonly used in Named Entity Recognition (NER) tasks.

However, each student has a sequence of interactions with associated labels in KT. Therefore, the LKT task is a blend of sequence classification and token classification. Formally, for a student $i$, the interaction sequence is represented as $\mathbf{x}_i = ([CLS], \mathbf{c}_i^1, \mathbf{q}_i^1, \mathbf{r}_i^1, \dots, \mathbf{c}_i^T, \mathbf{q}_i^T, \mathbf{r}_i^T, [EOS])$ where $\mathbf{c}_i^t$, $\mathbf{q}_i^t$, and $\mathbf{r}_i^t$ are the text representations of the KC, question, and response, respectively, at time $t$ for the student $i$.

Each student's sequence consists of a combination of KC text $\mathbf{c}_i$, question text $\mathbf{q}_i$, and student answers $\mathbf{r}_i$. The entire text is concatenated into a single line. After each KC text $\mathbf{c}_i$ and question text $\mathbf{q}_i$, if the student answers $\mathbf{r}_i$ correctly, the $\mathbf{r}_i$ is special token $[CORRECT]$; otherwise, the $\mathbf{r}_i$ is $[INCORRECT]$. However, 15\% of the $[CORRECT]$ or $[INCORRECT]$ tokens are replaced with $[MASK]$ tokens, inspired by BERT \cite{devlin2018bert}.

Encoder-based PLMs are fine-tuned on the formatted dataset to create the LKT model. Specifically, the PLMs are trained to predict the probability of the $[MASK]$ token, which ranges from 0 to 1. The PLMs $\mathcal{L}_{\theta_{pre}}$ takes $\mathbf{x}_i$ as input and generates logits for each token, and the logits of the $[MASK]$ token at position $m$ is extracted as:

\begin{equation}
\mathbf{H}_i = \mathcal{L}_{\theta_{pre}}(\mathbf{x}_i)
\end{equation}

\begin{equation}
\mathbf{h}_i^m = \mathbf{H}_i[m]
\end{equation}

Finally, the probability of correctness at the $[MASK]$ position is predicted using a sigmoid $\sigma$ function:

\begin{equation}
\hat{y}_i^m = \sigma(\mathbf{h}_i^m)
\end{equation}

After fine-tuning, the LKT model can predict the probability of correctness at the $[MASK]$ position. This simple yet effective approach outperforms previous KT methods. The model is trained using a binary cross-entropy loss between the predicted probability $\hat{y}_i^m$ and the actual correctness $y_i^m \in \{0, 1\}$, where $N$ is the total number of students:

\begin{equation}
\mathcal{L} = -\frac{1}{N}\sum_{i=1}^N [y_i^m \log(\hat{y}_i^m) + (1 - y_i^m) \log(1 - \hat{y}_i^m)]
\end{equation}

Figure \ref{fig:whole_arch} is comparison of DKT (\textit{Left}) and LKT framework (\textit{Right}). Note that DKTs have many variants, the figure is simplified for clarity

%% file: 4_result.tex
% Please add the following required packages to your document preamble:
\begin{table*}[hbt!]
\small
\centering
\resizebox{0.9\textwidth}{!}{%
\begin{tabular}{clcccc}
\toprule
\multirow{2}{*}{Type} & \multicolumn{1}{c}{\multirow{2}{*}{Models}} & \multicolumn{2}{c}{DBE-KT22}  & \multicolumn{2}{c}{XES3G5M-T}   \\
                      & \multicolumn{1}{c}{}                        & AUC           & ACC           & AUC           & ACC           \\
\toprule
LKT                   & BERT                                        & 0.7452±0.0058 & 0.7769±0.0100 & 0.8458±0.0011 & 0.8390±0.0015 \\
LKT                   & ALBERT                                      & 0.6911±0.0076 & 0.7617±0.0062 & 0.8252±0.0302 & 0.8318±0.0130 \\
LKT                   & DistilBERT                                  & 0.7593±0.0028 & 0.7836±0.0032 & 0.8440±0.0026 & 0.8364±0.0056 \\
LKT                   & RoBERTa                                     & 0.7643±0.0099 & 0.7673±0.0056 & \underline{0.8508±0.0021} & \underline{0.8420±0.0017} \\
LKT                   & ELECTRA                                     & 0.7250±0.0374 & 0.7751±0.0158 & 0.8468±0.0036 & 0.8419±0.0018 \\
LKT                   & ERNIE-2.0                                   & 0.7633±0.0084 & 0.7635±0.0051 & 0.8480±0.0036 & 0.8407±0.0015 \\
LKT                   & DeBERTa-v3                                  & 0.7415±0.0438 & 0.7781±0.0096 & \textbf{0.8513±0.0032} & \textbf{0.8421±0.0010} \\
DKT                   & DKT                                         & 0.7819±0.0040 & 0.7902±0.0070 & 0.7852±0.0006 & 0.8173±0.0002 \\
DKT                   & DKVMN                                       & \underline{0.7831±0.0049} & \underline{0.7926±0.0063} & 0.7792±0.0004 & 0.8155±0.0001 \\
DKT                   & SAKT                                        & 0.7782±0.0034 & 0.7903±0.0060 & 0.7693±0.0008 & 0.8124±0.0002 \\
DKT                   & GKT (PAM)                                   & 0.7307±0.0469 & 0.7283±0.0466 & 0.7727±0.0006 & 0.8135±0.0004 \\
DKT                   & AKT                                         & \textbf{0.7984±0.0037} & \textbf{0.7953±0.0073} & 0.8207±0.0008 & 0.8273±0.0007 
\\
\hline
\end{tabular}
}
\caption{
Performance of LKTs and DKTs. The performance metrics are AUC and ACC. The models are trained on 200 epochs and ten early stop settings. The performance of DKT in XES3G5M-T referenced \cite{liu2024xes3g5m}. In a big dataset (XES3G5M-T), LKTs have higher performance than DKTs, both AUC and ACC. The DeBERTa-V3 shows the best performance. However, in the small dataset (DBE-KT22), DKTs are better than LKTs.
}
\label{tb:performance}
\end{table*}

\section{Experimental Results}

\subsection{Experiment Setup}

\subsubsection{Models}

In this study, we employed two distinct types of models: LKT and DKT. The LKT models comprise bidirectional PLMs. We selected well-known PLMs from Hugging Face Transformers \cite{wolf-etal-2020-transformers}, including BERT \cite{devlin2018bert}, ALBERT \cite{lan2019albert}, DistilBERT \cite{sanh2019distilbert}, RoBERTa \cite{liu2019roberta}, ELECTRA-discriminator \cite{clark2020electra}, ERNIE-2.0-en \cite{sun2020ernie} and DeBERTa-v3 \cite{he2021debertav3}. For the experiments, we used the base size model of each LM (Table \ref{tb:performance})

For the DKT models, we selected DKT \cite{piech2015deep}, DKVMN \cite{zhang2017dynamic}, GKT \cite{nakagawa2019graph}, SAKT \cite{pandey2019self} and AKT \cite{ghosh2020context}. These models were chosen based on the criteria that the KT models use question IDs, knowledge concept IDs, and student responses as their primary features.

\subsubsection{Datasets}

The DBE-KT22 \cite{abdelrahman2022dbe} and XES3G5M \cite{liu2024xes3g5m} datasets contribute significant textual features regarding questions and KCs within the field of KT (\href{https://dataverse.ada.edu.au/dataset.xhtml?persistentId=doi:10.26193/6DZWOH}{DBE-KT-22}). The DBE-KT22 \cite{abdelrahman2022dbe} dataset includes exercise information from undergraduate students participating in a Relational Databases course at the Australian National University from 2018 to 2021, gathered via the CodeBench platform. This dataset is notable for its wide-ranging content, featuring contributions from students of various fields of study.

On the other hand, the XES3G5M dataset \cite{liu2024xes3g5m}, created by the TAL Education Group, covers the academic performance of third-grade students in mathematics, recording over 5 million interactions (\href{https://github.com/ai4ed/XES3G5M?tab=readme-ov-file}{XES3G5M}). This dataset offers insights from the activities of more than 18,000 students answering around 8,000 math questions, where the original textual data is in Chinese. Given the predominance of English in the training of most PLMs, the entire textual content of XES3G5M was translated into English using GPT-4-turbo \cite{achiam2023gpt}, resulting in the XES3G5M-T.

Utilizing the BERT-base tokenizer, the DBE-KT22 dataset amounts to 13 million tokens, while the XES3G5M-T encompasses 271 million tokens. Table \ref{tb:dataset} presents detailed information about these datasets.

\begin{table}[]
\centering
\resizebox{0.9\columnwidth}{!}{%
\begin{tabular}{lll}
\toprule
& DBE-KT22 & XES3G5M-T \\
\midrule
\#Students      & 1,361            & 18,066 \\
\#KCs           & 98               & 865 \\
\#Questions     & 212              & 7,652 \\
\#Interactions  & 167,222          & 5,549,635 \\
Language        & English          & English (Translated) \\
\#Tokens        & 13M              & 271M \\
\bottomrule
\end{tabular}
}
\caption{
Dataset information about DBE-KT22 and XES3G5M-T. Note that \#Tokens are counted after making the training data format for LKT.
}
\label{tb:dataset}
\end{table}

\subsubsection{Training and Evaluation}

We conducted standard five-fold cross-validation for all models and datasets to evaluate the model's performance. The number of epochs for training is set to 200, and the early stopping threshold, activated when the validation loss does not improve, is ten epochs. The maximum sequence length is 512, and the batch size is 512. Due to limited resources, we employed the gradient accumulation technique to train our model with the desired large batch size.

% 여기에는 성능 비교 그래프를 넣기
\begin{figure*}[hbt!]
\centering
\includegraphics[width=1\linewidth]{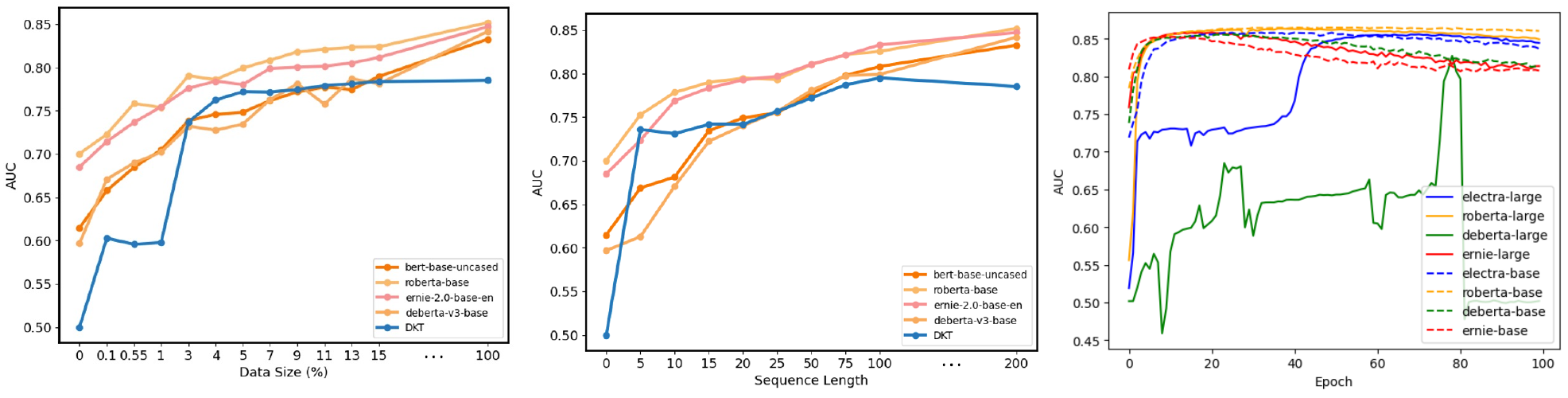}
\caption{
We examine the cold start problem in KT, which changes performance as model size increases. The \textit{Left} shows the AUC of LKTs pre-trained on DBE-KT22 and DKT trained only on XES3G5M-T across different data sizes (0.1\%, 0.5\%, 1\%, 3\%, ..., 15\%). The LKTs demonstrate robustness to the cold start problem. The \textit{Center} displays AUC scores for different sequence lengths per student (5, 10, 20, etc.). The RoBERTa-based LKT performs well with fewer data, indicating initial solid performance. The \textit{Right} compares the performance of large and base LKTs. Solid lines represent large models, while dashed lines represent base models. RoBERTa and ERNIE models maintain stable AUC performance regardless of size.
}
\label{fig:scail_and_cold_v3}
\end{figure*}

\subsection{Performance}

Table \ref{tb:performance} presents the performance of various LKT and DKT models. The evaluation metrics used are the Area Under the ROC Curve (AUC) and accuracy (ACC).

In the large data set (XES3G5M-T), LKT models generally outperform DKT models in both AUC and ACC. Among the LKT models, DeBERTa-V3 shows the best performance with an AUC of 0.8513±0.0032 and an ACC of 0.8421±0.0010. RoBERTa also demonstrates strong results, with an AUC of 0.8508±0.0021 and an ACC of 0.8420±0.0017. These findings suggest that LKT models effectively capture student knowledge states in large-scale datasets.

On the smaller dataset (DBE-KT22), DKT models exhibit better performance than LKT models. The AKT model achieves the highest AUC of 0.7984±0.0037 and ACC of 0.7953±0.0073, followed closely by the DKVMN model with an AUC of 0.7831±0.0049 and an ACC of 0.7926±0.0063. This indicates that DKT models may be more suitable for KT tasks when the available dataset is small.

\begin{table*}[hbt!]
\centering
\begin{tabular}{l}
\hline
\textbf{(a) Mean Attention Scores of LKT-BERT:} \\
\toprule
\textit{Concept}: Cardinality\\
\textit{Question}: The cardinality of a set is the number of elements of the set. \underline{What is the cardinality} \\ \underline{of the set of odd positive integers less than 10?}\\
\textit{Response}: \underline{\lbrack INCORRECT \rbrack}\\
\includegraphics[width=1\linewidth]{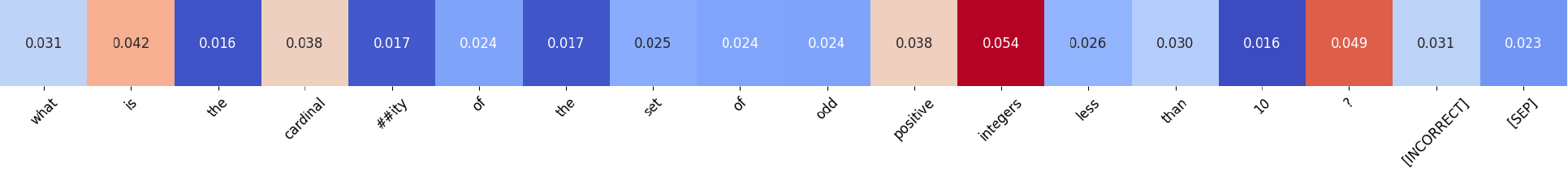} \\
\hline
\textbf{(b) LIME Analysis Result:} \\
\toprule
\includegraphics[width=1\linewidth]{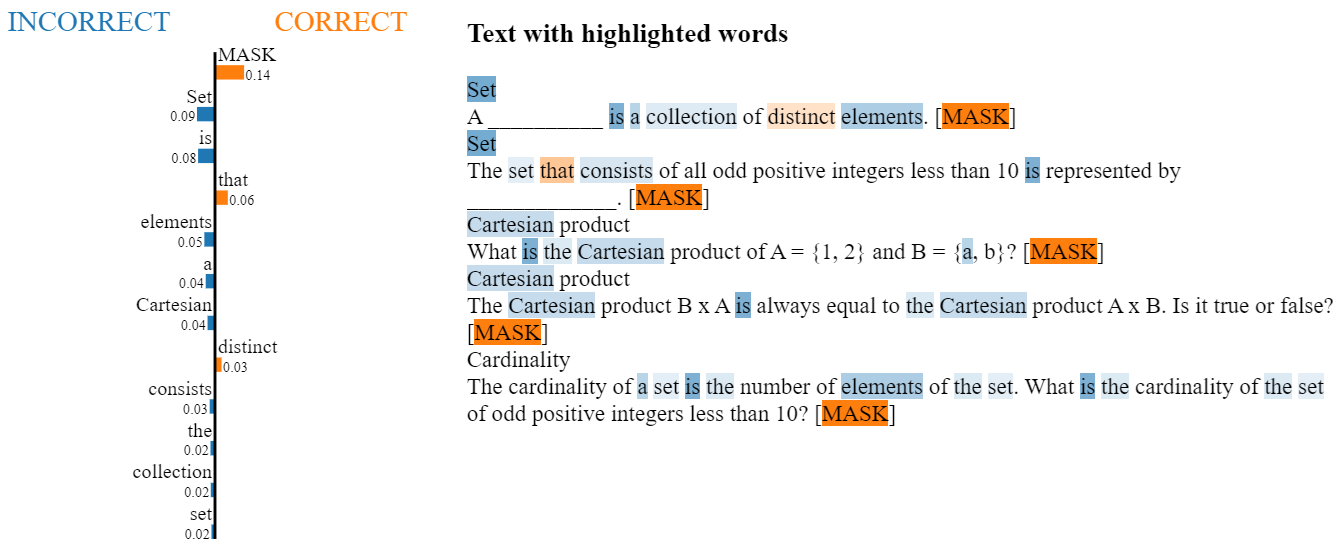} \\
\hline
\end{tabular}
\caption{
\textbf{(a)} shows the mean attention scores of the first head and layer of the LKT-BERT model. The figure shows higher averaged attention scores for the word `integers.' These averaged attention scores provide insights into which tokens the model focuses on during the attention mechanism. \textbf{(b)} demonstrates the LIME analysis result of the model. The highlighted words indicate that the model effectively focuses on the words in each test item related to its associated concept.
}
\label{tab:attention_1d_and_lime}
\end{table*}

\subsection{Cold Start Problem}

% cold start zero-shot 결과
\begin{figure}[hbt]
\centering
\includegraphics[width=0.9\linewidth]{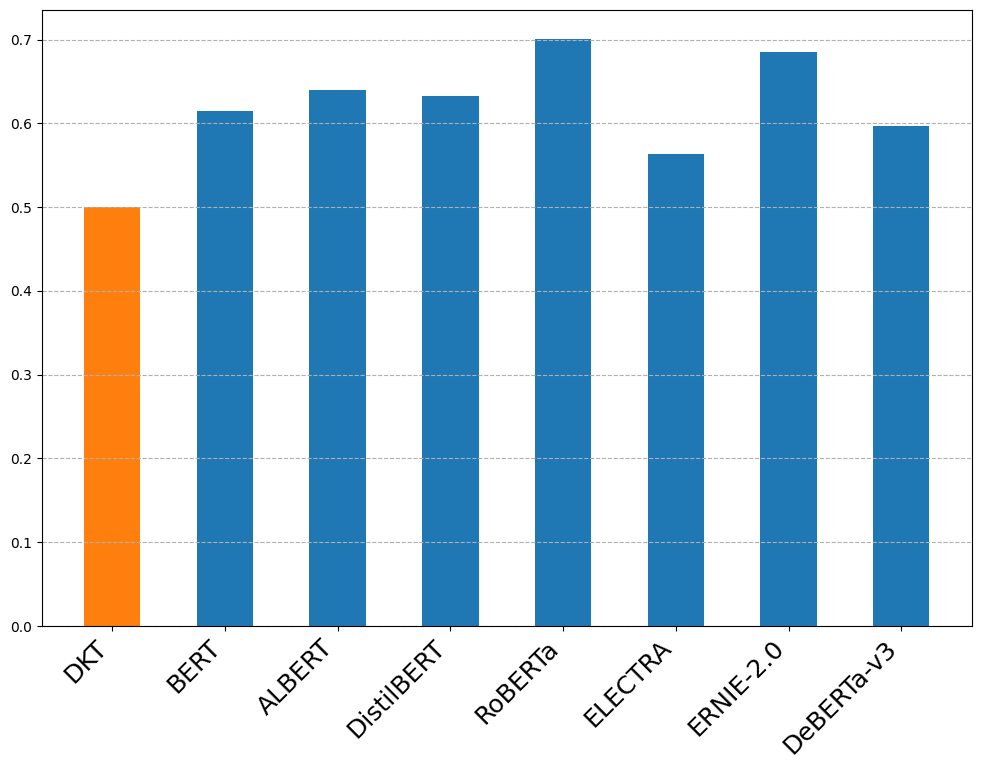}
\caption{
Performance comparison (AUC) of DKT and LKT models on XES3G5M-T data. The LKT model, pre-trained on DBE-KT22, outperformed the DKT model without additional training on new data. Note that DKT's performance is 0.5 due to its inability to utilize pre-training.
}
\label{fig:cold_zero}
\end{figure}

The cold start problem is a common challenge in KT, where models must predict performance with insufficient data. To address this issue, we compared its performance with two scenarios based on previous KT research \cite{zhao2020cold, zhang2021knowledge, slater2018degree}: (1) when the overall amount of data is limited, and (2) when the sequence length for each student is short.

In the first setting (Figure \ref{fig:scail_and_cold_v3} \textit{Left} and Figure \ref{fig:init_cold}), we pre-trained LKT on the DBE-KT22 dataset and applied it directly to the small XES3G5M-T dataset. For comparison, we trained DKT only on the XES3G5M-T dataset. The results show that LKT outperforms DKT on the small dataset. Note that Figure \ref{fig:init_cold} shows the results of this experiment for the RoBERTa model, pre-trained on DBE-KT22, and DKT.

In the second setting (Figure \ref{fig:scail_and_cold_v3}, \textit{Center}), we examined the performance of models across different sequence lengths per student (5, 10, 20, etc.). The LKT models, pre-trained on DBE-KT22, showed that the RoBERTa-based LKT performs well even with less data, indicating initial solid performance. These results underscore LKT's capability to leverage pre-trained knowledge and textual features to enhance performance in cold start conditions.

Moreover, we investigated the zero-shot performance of LKTs (Figure \ref{fig:cold_zero}). The performance comparison of DKT and LKTs on the XES3G5M-T dataset reveals that the LKTs, pre-trained on DBE-KT22, outperformed the DKT model in a zero-shot scenario. The DKT model had an AUC of 0.5 because it wasn't measured due to its reliance on domain-specific numerical data, making pre-training complex \cite{liu2021improving}. In contrast, the LKT model benefitted from pre-training, demonstrating the effectiveness of this approach in enhancing model performance on new datasets even without additional training.

Overall, these findings highlight the robustness of LKT models in handling the cold start problem. Future research should explore leveraging pre-trained models for other KT tasks further.

\subsection{The Impact of Language Model Size on Performance}

Figure \ref{fig:scail_and_cold_v3}, \textit{Right} compares the performance of large and base LKTs, with solid lines representing large models and dashed lines representing base models. To examine the relationship between PLM size and performance, we fine-tuned four PLMs: RoBERTa-Large (335M), Electra-Large (335M), ERNIE-2.0-en-Large (335M), and DeBERTa-v3-Large (435M). The experiments showed that large PLMs require large data, warm-up steps, and longer epochs for optimal performance. We used the XES3G5M-T dataset, 2,000 warm-up steps, and 100 epochs without early stopping.

During training, larger models were challenging to train, but their performance improved with warm-up steps. Approximately 2,000 warm-up steps were introduced to facilitate training and enhance AUC performance.

Interestingly, RoBERTa and ERNIE showed rapid performance improvements early in training, while Electra and DeBERTa exhibited significant performance boosts at specific points during training. Despite these differences, all four models achieved a maximum AUC performance of over 0.8, demonstrating their effectiveness in the given task.

\subsection{Which tokens are important in LKT?}

While DKTs only utilize the sequence of question and concept numbers, making it challenging to interpret which parts of the sequence influence the model's performance, LKTs incorporate textual features of questions and concepts, enabling interpretation by analyzing the sequences' tokens. We employed attention maps and local interpretable model-agnostic explanations (LIME) \cite{ribeiro2016should} to investigate which tokens significantly impact the model's prediction results. 

Table \ref{tab:attention_1d_and_lime}-(a) illustrates the 1D attention map of the first head and layer of the LKT-BERT model, representing the mean attention scores. The attention score for the word `integers' is notably higher than the scores for other words, suggesting that the presence of `integers' influences the model's prediction results. However, it is essential to note that while higher attention scores indicate the model's focus on specific tokens, they do not necessarily directly impact the final prediction, as the model's output is influenced by the complex interaction of attention across multiple layers and heads.

Table \ref{tab:attention_1d_and_lime}-(b) illustrates the interpretation of model prediction using the LIME technique. The results show a list of words and their corresponding weights, indicating how much each word contributes to the model's prediction of $[CORRECT]$ or $[INCORRECT]$. The text with highlighted words illustrates which words in each test item contribute to predictions of correctness (highlighted in orange) or incorrectness (highlighted in blue). The highlighted words indicate that the model effectively focuses on the words in each test item related to its associated concept, which appropriately influences the model's predictions. For instance, the words that consist of the test item related to the concept 'set' were 'is,' 'a', 'collection,' 'distinct,' and 'elements.' In addition, the highlighted words indicate that the model effectively focuses on words closely related to the matching concepts for prediction. Note that the input data sequence consisted of a concept, a corresponding test item, and masked answers. This data was part of the sequence of test items one of the learners received and solved.

\subsection{In-depth Analysis of Embedding}

\begin{figure}[hbt]
\centering
\includegraphics[width=1\linewidth]{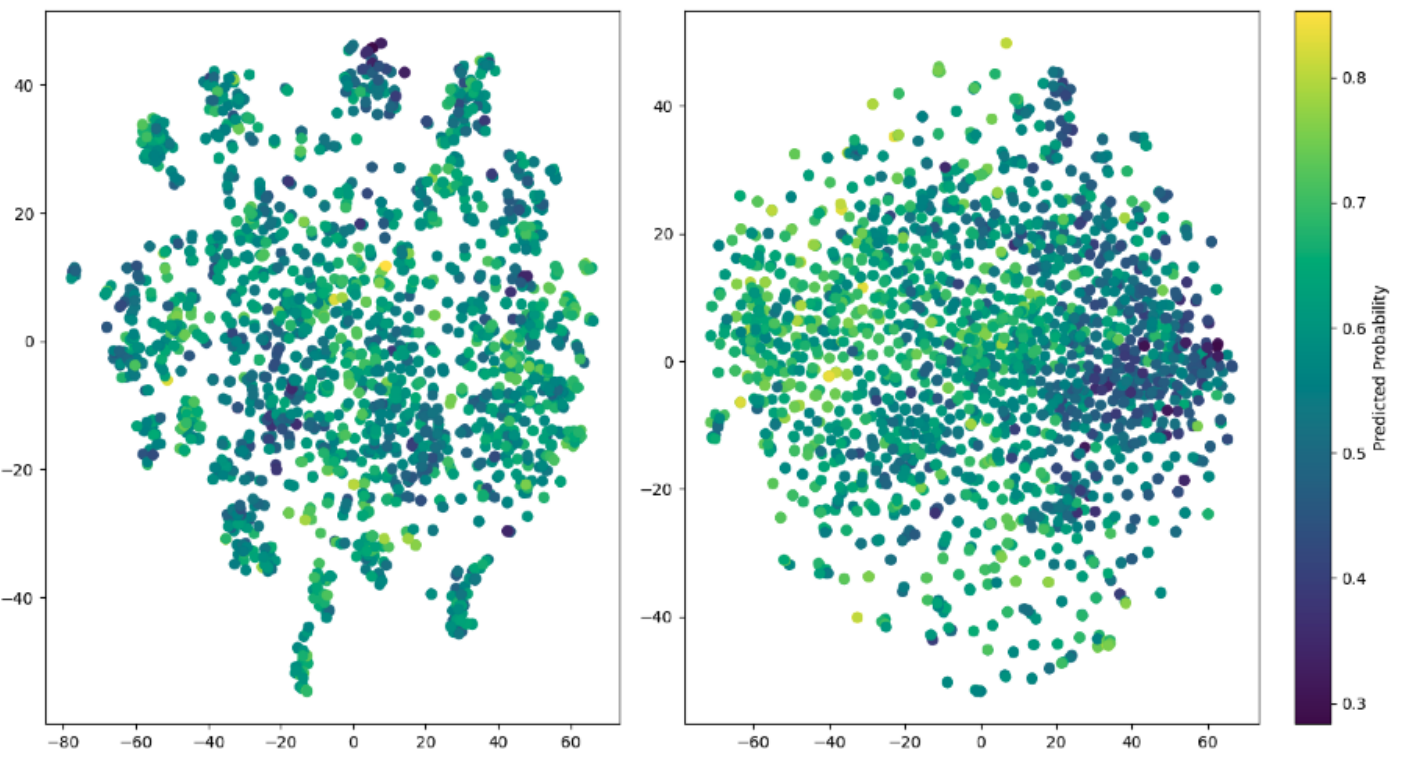}
\caption{
Visualization of the embedding vector with T-SNE. \textit{Left} shows BERT and \textit{Right} shows the result of BERT-LKT embedding. We can see that the results of BERT-LKT embedding represent the correctness probability well.
}
\label{fig:embedding}
\end{figure}

We visualized the embedding vectors of the BERT and BERT-LKT models using T-SNE \cite{van2008visualizing}. Figure \ref{fig:embedding} shows the embeddings of the BERT model on the \textit{Left} and the embeddings of the BERT-LKT model on the \textit{Right}.

The BERT model's embeddings are randomly distributed, indicating it does not effectively capture the probability of correct answers. In contrast, the BERT-LKT model's embeddings form distinct clusters based on the probability of correct answers, with high probabilities grouped on the \textit{Right} and lower probabilities moving towards the \textit{Left}.

These results highlight the BERT-LKT model's superior ability to encode educational data and reflect students' performance probabilities, demonstrating the benefits of integrating KT into the BERT model.

%% file: 5_conclusion.tex
\section{Conclusion}

In this research, we proposed a novel framework that integrates encoder-based PLMs with KT. Leveraging the rich semantic representations captured by PLMs, our LKT framework outperforms state-of-the-art KT models on larger datasets, enabling accurate predictions of student performance on new questions and concepts even with limited data. This study also explores the impact of language model size on performance, showing that larger models can achieve higher AUC scores with appropriate training strategies. Additionally, the LKT framework addresses the cold-start problem in KT using the semantic knowledge captured by PLMs.

Our LKT framework provides insights into which parts of knowledge concepts and questions affect student performance, aiding in developing more effective educational materials and interventions. This research highlights the potential of integrating PLMs with KT, and opening new avenues for future research. The study contributes to creating refined personalized learning paths and improving feedback mechanisms to address students' misconceptions.

\section{Limitation}

While our LKT framework has demonstrated state-of-the-art performance on large KT benchmark datasets, a few limitations warrant further investigation and improvement. 

First, additional analyses are necessary to fully understand which aspects of the PLMs are responsible for LKT's success and what specific understanding of the question and concept text the PLMs capture. Future work should explore techniques to improve the interpretability of LKT models, such as attention visualization and probing tasks.

Second, only some KT datasets containing textual features are publicly available, limiting the ability to validate the effectiveness of the LKT framework on a broader range of KT tasks.

Finally, our LKT framework incorporates textual features from questions and concepts. However, educational data often includes other modalities, such as images, videos, and interactive elements. Extending LKT to handle multi-modal input could lead to further performance improvements and a more comprehensive understanding of student knowledge. Addressing these limitations will help refine the LKT framework and pave the way for more effective and interpretable KT models to support personalized learning experiences better.

\section{Ethical Consideration}

This study utilizes two datasets: DBE-KT-22 and XES3G5M. To protect personal identifying information in these datasets, we preprocessed all personal identification data, ensuring that only anonymized data remained for research purposes.

Regarding license information, the DBE-KT-22 dataset can be used solely for analytical purposes, as detailed in the \href{https://dataverse.ada.edu.au/dataset.xhtml?persistentId=doi:10.26193/6DZWOH&version=1.0&selectTab=termsTab}{DBE-KT-22 License Information}. In contrast, the XES3G5M dataset is available under the MIT license, allowing for both research and commercial use, as specified in the \href{https://github.com/ai4ed/XES3G5M/blob/main/LICENSE}{XES3G5M License Information}. Therefore, future researchers and companies must consider the respective licenses when utilizing these datasets.

Additionally, we employed ChatGPT (GPT-4) to paraphrase and enhance the fluency of the writing in this paper.

%% file: main.bbl
\begin{thebibliography}{41}
\expandafter\ifx\csname natexlab\endcsname\relax\def\natexlab#1{#1}\fi

\bibitem[{Abdelrahman et~al.(2022)Abdelrahman, Abdelfattah, Wang, and Lin}]{abdelrahman2022dbe}
Ghodai Abdelrahman, Sherif Abdelfattah, Qing Wang, and Yu~Lin. 2022.
\newblock Dbe-kt22: A knowledge tracing dataset based on online student evaluation.
\newblock \emph{arXiv preprint arXiv:2208.12651}.

\bibitem[{Abdelrahman et~al.(2023)Abdelrahman, Wang, and Nunes}]{abdelrahman2023knowledge}
Ghodai Abdelrahman, Qing Wang, and Bernardo Nunes. 2023.
\newblock Knowledge tracing: A survey.
\newblock \emph{ACM Computing Surveys}, 55(11):1--37.

\bibitem[{Achiam et~al.(2023)Achiam, Adler, Agarwal, Ahmad, Akkaya, Aleman, Almeida, Altenschmidt, Altman, Anadkat et~al.}]{achiam2023gpt}
Josh Achiam, Steven Adler, Sandhini Agarwal, Lama Ahmad, Ilge Akkaya, Florencia~Leoni Aleman, Diogo Almeida, Janko Altenschmidt, Sam Altman, Shyamal Anadkat, et~al. 2023.
\newblock Gpt-4 technical report.
\newblock \emph{arXiv preprint arXiv:2303.08774}.

\bibitem[{Ayu(2020)}]{ayu2020online}
Mutiara Ayu. 2020.
\newblock Online learning: Leading e-learning at higher education.
\newblock \emph{The Journal of English Literacy Education: The Teaching and Learning of English as a Foreign Language}, 7(1):47--54.

\bibitem[{Clark et~al.(2020)Clark, Luong, Le, and Manning}]{clark2020electra}
Kevin Clark, Minh-Thang Luong, Quoc~V Le, and Christopher~D Manning. 2020.
\newblock Electra: Pre-training text encoders as discriminators rather than generators.
\newblock \emph{arXiv preprint arXiv:2003.10555}.

\bibitem[{Corbett and Anderson(1994)}]{corbett1994knowledge}
Albert~T Corbett and John~R Anderson. 1994.
\newblock Knowledge tracing: Modeling the acquisition of procedural knowledge.
\newblock \emph{User modeling and user-adapted interaction}, 4:253--278.

\bibitem[{Das et~al.(2021)Das, Zhang, Baker, and Scruggs}]{das2021new}
Rohini Das, Jiayi Zhang, Ryan~S Baker, and Richard Scruggs. 2021.
\newblock A new interpretation of knowledge tracing models' predictive performance in terms of the cold start problem.
\newblock In \emph{EDM (Workshops)}.

\bibitem[{Devlin et~al.(2018)Devlin, Chang, Lee, and Toutanova}]{devlin2018bert}
Jacob Devlin, Ming-Wei Chang, Kenton Lee, and Kristina Toutanova. 2018.
\newblock Bert: Pre-training of deep bidirectional transformers for language understanding.
\newblock \emph{arXiv preprint arXiv:1810.04805}.

\bibitem[{Dhawan(2020)}]{dhawan2020online}
Shivangi Dhawan. 2020.
\newblock Online learning: A panacea in the time of covid-19 crisis.
\newblock \emph{Journal of educational technology systems}, 49(1):5--22.

\bibitem[{Ghosh et~al.(2020)Ghosh, Heffernan, and Lan}]{ghosh2020context}
Aritra Ghosh, Neil Heffernan, and Andrew~S Lan. 2020.
\newblock Context-aware attentive knowledge tracing.
\newblock In \emph{Proceedings of the 26th ACM SIGKDD international conference on knowledge discovery \& data mining}, pages 2330--2339.

\bibitem[{Gikandi et~al.(2011)Gikandi, Morrow, and Davis}]{gikandi2011online}
Joyce~Wangui Gikandi, Donna Morrow, and Niki~E Davis. 2011.
\newblock Online formative assessment in higher education: A review of the literature.
\newblock \emph{Computers \& education}, 57(4):2333--2351.

\bibitem[{He et~al.(2021)He, Gao, and Chen}]{he2021debertav3}
Pengcheng He, Jianfeng Gao, and Weizhu Chen. 2021.
\newblock Debertav3: Improving deberta using electra-style pre-training with gradient-disentangled embedding sharing.
\newblock \emph{arXiv preprint arXiv:2111.09543}.

\bibitem[{Jung et~al.(2023)Jung, Yoo, Yoon, and Jang}]{jung2023language}
Heeseok Jung, Jaesang Yoo, Yohaan Yoon, and Yeonju Jang. 2023.
\newblock Language proficiency enhanced knowledge tracing.
\newblock In \emph{International Conference on Intelligent Tutoring Systems}, pages 3--15. Springer.

\bibitem[{Khurana et~al.(2023)Khurana, Koli, Khatter, and Singh}]{khurana2023natural}
Diksha Khurana, Aditya Koli, Kiran Khatter, and Sukhdev Singh. 2023.
\newblock Natural language processing: State of the art, current trends and challenges.
\newblock \emph{Multimedia tools and applications}, 82(3):3713--3744.

\bibitem[{Lan et~al.(2019)Lan, Chen, Goodman, Gimpel, Sharma, and Soricut}]{lan2019albert}
Zhenzhong Lan, Mingda Chen, Sebastian Goodman, Kevin Gimpel, Piyush Sharma, and Radu Soricut. 2019.
\newblock Albert: A lite bert for self-supervised learning of language representations.
\newblock \emph{arXiv preprint arXiv:1909.11942}.

\bibitem[{Lee et~al.(2022{\natexlab{a}})Lee, Park, Kim, Choi, and Kim}]{lee2022monacobert}
Unggi Lee, Yonghyun Park, Yujin Kim, Seongyune Choi, and Hyeoncheol Kim. 2022{\natexlab{a}}.
\newblock Monacobert: Monotonic attention based convbert for knowledge tracing.
\newblock \emph{arXiv preprint arXiv:2208.12615}.

\bibitem[{Lee et~al.(2022{\natexlab{b}})Lee, Chun, Lee, Park, and Park}]{lee2022contrastive}
Wonsung Lee, Jaeyoon Chun, Youngmin Lee, Kyoungsoo Park, and Sungrae Park. 2022{\natexlab{b}}.
\newblock Contrastive learning for knowledge tracing.
\newblock In \emph{Proceedings of the ACM Web Conference 2022}, pages 2330--2338.

\bibitem[{Leo et~al.(2021)Leo, Alsharari, Abbas, and Alshurideh}]{leo2021offline}
Shirley Leo, Nizar~Mohammad Alsharari, Jainambu Abbas, and Muhammad~Turki Alshurideh. 2021.
\newblock From offline to online learning: A qualitative study of challenges and opportunities as a response to the covid-19 pandemic in the uae higher education context.
\newblock \emph{The effect of coronavirus disease (COVID-19) on business intelligence}, pages 203--217.

\bibitem[{Li et~al.(2024)Li, Yu, Ouyang, Liu, Rong, Li, and Xiong}]{li2024explainable}
Haoxuan Li, Jifan Yu, Yuanxin Ouyang, Zhuang Liu, Wenge Rong, Juanzi Li, and Zhang Xiong. 2024.
\newblock Explainable few-shot knowledge tracing.
\newblock \emph{arXiv preprint arXiv:2405.14391}.

\bibitem[{Liu et~al.(2019{\natexlab{a}})Liu, Huang, Yin, Chen, Xiong, Su, and Hu}]{liu2019ekt}
Qi~Liu, Zhenya Huang, Yu~Yin, Enhong Chen, Hui Xiong, Yu~Su, and Guoping Hu. 2019{\natexlab{a}}.
\newblock Ekt: Exercise-aware knowledge tracing for student performance prediction.
\newblock \emph{IEEE Transactions on Knowledge and Data Engineering}, 33(1):100--115.

\bibitem[{Liu et~al.(2019{\natexlab{b}})Liu, Ott, Goyal, Du, Joshi, Chen, Levy, Lewis, Zettlemoyer, and Stoyanov}]{liu2019roberta}
Yinhan Liu, Myle Ott, Naman Goyal, Jingfei Du, Mandar Joshi, Danqi Chen, Omer Levy, Mike Lewis, Luke Zettlemoyer, and Veselin Stoyanov. 2019{\natexlab{b}}.
\newblock Roberta: A robustly optimized bert pretraining approach.
\newblock \emph{arXiv preprint arXiv:1907.11692}.

\bibitem[{Liu et~al.(2021)Liu, Yang, Chen, Shen, Zhang, and Yu}]{liu2021improving}
Yunfei Liu, Yang Yang, Xianyu Chen, Jian Shen, Haifeng Zhang, and Yong Yu. 2021.
\newblock Improving knowledge tracing via pre-training question embeddings.
\newblock In \emph{Proceedings of the Twenty-Ninth International Conference on International Joint Conferences on Artificial Intelligence}, pages 1556--1562.

\bibitem[{Liu et~al.(2024)Liu, Liu, Guo, Chen, Huang, Zhao, Tang, Luo, and Weng}]{liu2024xes3g5m}
Zitao Liu, Qiongqiong Liu, Teng Guo, Jiahao Chen, Shuyan Huang, Xiangyu Zhao, Jiliang Tang, Weiqi Luo, and Jian Weng. 2024.
\newblock Xes3g5m: A knowledge tracing benchmark dataset with auxiliary information.
\newblock \emph{Advances in Neural Information Processing Systems}, 36.

\bibitem[{Lundberg and Lee(2017)}]{lundberg2017unified}
Scott~M Lundberg and Su-In Lee. 2017.
\newblock A unified approach to interpreting model predictions.
\newblock \emph{Advances in neural information processing systems}, 30.

\bibitem[{Nakagawa et~al.(2019)Nakagawa, Iwasawa, and Matsuo}]{nakagawa2019graph}
Hiromi Nakagawa, Yusuke Iwasawa, and Yutaka Matsuo. 2019.
\newblock Graph-based knowledge tracing: modeling student proficiency using graph neural network.
\newblock In \emph{IEEE/WIC/ACM International Conference on Web Intelligence}, pages 156--163.

\bibitem[{Ogange et~al.(2018)Ogange, Agak, Okelo, and Kiprotich}]{ogange2018student}
Betty~Obura Ogange, John~O Agak, Kevin~Odhiambo Okelo, and Peter Kiprotich. 2018.
\newblock Student perceptions of the effectiveness of formative assessment in an online learning environment.
\newblock \emph{Open Praxis}, 10(1):29--39.

\bibitem[{Pandey and Karypis(2019)}]{pandey2019self}
Shalini Pandey and George Karypis. 2019.
\newblock A self-attentive model for knowledge tracing.
\newblock \emph{arXiv preprint arXiv:1907.06837}.

\bibitem[{Piech et~al.(2015)Piech, Bassen, Huang, Ganguli, Sahami, Guibas, and Sohl-Dickstein}]{piech2015deep}
Chris Piech, Jonathan Bassen, Jonathan Huang, Surya Ganguli, Mehran Sahami, Leonidas~J Guibas, and Jascha Sohl-Dickstein. 2015.
\newblock Deep knowledge tracing.
\newblock \emph{Advances in neural information processing systems}, 28.

\bibitem[{Ribeiro et~al.(2016)Ribeiro, Singh, and Guestrin}]{ribeiro2016should}
Marco~Tulio Ribeiro, Sameer Singh, and Carlos Guestrin. 2016.
\newblock " why should i trust you?" explaining the predictions of any classifier.
\newblock In \emph{Proceedings of the 22nd ACM SIGKDD international conference on knowledge discovery and data mining}, pages 1135--1144.

\bibitem[{Sanh et~al.(2019)Sanh, Debut, Chaumond, and Wolf}]{sanh2019distilbert}
Victor Sanh, Lysandre Debut, Julien Chaumond, and Thomas Wolf. 2019.
\newblock Distilbert, a distilled version of bert: smaller, faster, cheaper and lighter.
\newblock \emph{arXiv preprint arXiv:1910.01108}.

\bibitem[{Shen et~al.(2024)Shen, Liu, Huang, Zheng, Yin, Wang, and Chen}]{shen2024survey}
Shuanghong Shen, Qi~Liu, Zhenya Huang, Yonghe Zheng, Minghao Yin, Minjuan Wang, and Enhong Chen. 2024.
\newblock A survey of knowledge tracing: Models, variants, and applications.
\newblock \emph{IEEE Transactions on Learning Technologies}.

\bibitem[{Slater and Baker(2018)}]{slater2018degree}
Stefan Slater and Ryan~S Baker. 2018.
\newblock Degree of error in bayesian knowledge tracing estimates from differences in sample sizes.
\newblock \emph{Behaviormetrika}, 45(2):475--493.

\bibitem[{Song et~al.(2022)Song, Li, Cai, Yang, Yang, and Liu}]{song2022survey}
Xiangyu Song, Jianxin Li, Taotao Cai, Shuiqiao Yang, Tingting Yang, and Chengfei Liu. 2022.
\newblock A survey on deep learning based knowledge tracing.
\newblock \emph{Knowledge-Based Systems}, 258:110036.

\bibitem[{Su et~al.(2018)Su, Liu, Liu, Huang, Yin, Chen, Ding, Wei, and Hu}]{su2018exercise}
Yu~Su, Qingwen Liu, Qi~Liu, Zhenya Huang, Yu~Yin, Enhong Chen, Chris Ding, Si~Wei, and Guoping Hu. 2018.
\newblock Exercise-enhanced sequential modeling for student performance prediction.
\newblock In \emph{Proceedings of the AAAI Conference on Artificial Intelligence}, volume~32.

\bibitem[{Sun et~al.(2020)Sun, Wang, Li, Feng, Tian, Wu, and Wang}]{sun2020ernie}
Yu~Sun, Shuohuan Wang, Yukun Li, Shikun Feng, Hao Tian, Hua Wu, and Haifeng Wang. 2020.
\newblock Ernie 2.0: A continual pre-training framework for language understanding.
\newblock In \emph{Proceedings of the AAAI conference on artificial intelligence}, volume~34, pages 8968--8975.

\bibitem[{Van~der Maaten and Hinton(2008)}]{van2008visualizing}
Laurens Van~der Maaten and Geoffrey Hinton. 2008.
\newblock Visualizing data using t-sne.
\newblock \emph{Journal of machine learning research}, 9(11).

\bibitem[{Wang et~al.(2024)Wang, Ma, Zhao, and Li}]{wang2024pre}
Wentao Wang, Huifang Ma, Yan Zhao, and Zhixin Li. 2024.
\newblock Pre-training question embeddings for improving knowledge tracing with self-supervised bi-graph co-contrastive learning.
\newblock \emph{ACM Transactions on Knowledge Discovery from Data}, 18(4):1--20.

\bibitem[{Wolf et~al.(2020)Wolf, Debut, Sanh, Chaumond, Delangue, Moi, Cistac, Rault, Louf, Funtowicz, Davison, Shleifer, von Platen, Ma, Jernite, Plu, Xu, Scao, Gugger, Drame, Lhoest, and Rush}]{wolf-etal-2020-transformers}
Thomas Wolf, Lysandre Debut, Victor Sanh, Julien Chaumond, Clement Delangue, Anthony Moi, Pierric Cistac, Tim Rault, Rémi Louf, Morgan Funtowicz, Joe Davison, Sam Shleifer, Patrick von Platen, Clara Ma, Yacine Jernite, Julien Plu, Canwen Xu, Teven~Le Scao, Sylvain Gugger, Mariama Drame, Quentin Lhoest, and Alexander~M. Rush. 2020.
\newblock \href {https://www.aclweb.org/anthology/2020.emnlp-demos.6} {Transformers: State-of-the-art natural language processing}.
\newblock In \emph{Proceedings of the 2020 Conference on Empirical Methods in Natural Language Processing: System Demonstrations}, pages 38--45, Online. Association for Computational Linguistics.

\bibitem[{Zhang et~al.(2017)Zhang, Shi, King, and Yeung}]{zhang2017dynamic}
Jiani Zhang, Xingjian Shi, Irwin King, and Dit-Yan Yeung. 2017.
\newblock Dynamic key-value memory networks for knowledge tracing.
\newblock In \emph{Proceedings of the 26th international conference on World Wide Web}, pages 765--774.

\bibitem[{Zhang et~al.(2021)Zhang, Das, Baker, and Scruggs}]{zhang2021knowledge}
Jiayi Zhang, Rohini Das, Ryan Baker, and Richard Scruggs. 2021.
\newblock Knowledge tracing models’ predictive performance when a student starts a skill.
\newblock In \emph{Proceedings of the 14th International Conference on Educational Data Mining. EDM, Paris, France}, pages 625--629.

\bibitem[{Zhao et~al.(2020)Zhao, Bhatt, Thille, Gattani, and Zimmaro}]{zhao2020cold}
Jinjin Zhao, Shreyansh Bhatt, Candace Thille, Neelesh Gattani, and Dawn Zimmaro. 2020.
\newblock Cold start knowledge tracing with attentive neural turing machine.
\newblock In \emph{Proceedings of the Seventh ACM Conference on Learning@ Scale}, pages 333--336.

\end{thebibliography}
